# Controlling Dialogue Generation with Semantic Exemplars

**Prakhar Gupta**[1], **Jeffrey P. Bigham**[1,2], **Yulia Tsvetkov**[1], **Amy Pavel**[2]

[1]Language Technologies Institute, Carnegie Mellon University,
[2]Human-Computer Interaction Institute, Carnegie Mellon University,
{prakharg,jbigham,ytsvetko,apavel}@cs.cmu.edu

## Abstract

Dialogue systems pretrained with large language models generate locally coherent responses, but lack the fine-grained control over responses necessary to achieve specific goals. A promising method for controlling generated responses is *exemplar-based generation*, in which models edit exemplar responses that are retrieved from training data, or hand-written to strategically address discourse-level goals, to fit new dialogue contexts. We present an **E**xemplar-based **D**ialogue **GE**neration model, **EDGE**, that uses the semantic frames present in exemplar responses to guide response generation. We show that controlling dialogue generation based on the semantic frames of exemplars improves the coherence of generated responses, while preserving semantic meaning and conversation goals present in exemplar responses.[1]

| Context | My friends and I have started eating vegan food since yesterday. |
|---|---|
| Exemplar Frames Responses | Eggs are very beneficial for your body. FOOD USEFULNESS BODY-PARTS Vegan food can be good for your health. Vegetables can do wonders for your body Vegan food is very healthy. |
| Exemplar Frames Responses | I want to drink milk as well. DESIRING INGESTION FOOD You want to eat some vegan food? We eat a lot of vegetables. It's delicious. We like to eat organic food. |

Table 1: EDGE generates responses to dialogue contexts by conditioning the response generation on the semantic frames of existing response exemplars to create coherent and controlled replies.

## 1 Introduction

Large pre-trained language models (Radford et al., 2019; Devlin et al., 2019) currently used to power dialogue generation systems produce increasingly fluent and appropriate responses for novel dialogue contexts (Wolf et al., 2019; Zhang et al., 2020; Budzianowski and Vulić, 2019). However, the generated responses are often uninformative or inconsistent with high-level constraints of a dialogue system and the tasks it supports. Prior work added high-level control for specific intents such as politeness (Niu and Bansal, 2018), emotions (Zhong et al., 2019) and persona (Song et al., 2019) through a fixed set of coarse labels, but these methods require manually labelling data for each new intent.

One approach for adding control over response intents is to use *response exemplars* that are hand-written or strategically curated to promote high-level goals without explicit labels. By conditioning on response exemplars, we can generate coherent responses that follow the intents of the exemplars without manually labeling vast amounts of data. Current exemplar-based methods (Cai et al., 2019b,a; Wu et al., 2019) have two key drawbacks: (1) the models often overfit to the training data, then produce incoherent responses by copying irrelevant tokens from exemplar responses into the generated responses, and (2) the models often learn to ignore the exemplars, then produce responses that are not controlled by the strategic exemplars.

To generate locally coherent responses that also adhere to high-level dialogue constraints, we present EDGE, a model that uses the semantic structure of an exemplar response, instead of the tokens of the exemplar response, to guide generation (Table 1). For a novel dialogue context, we retrieve a human-written response exemplar and represent it using its *semantic frames* (Fillmore, 1982). We then incorporate the dialogue context and the semantic frames of the response exemplars in a powerful pre-trained conditional language model (Radford et al., 2019), thereby combining the benefits of fluency of language models and the semantic guidance of the exemplar responses that are structured with rich linguistic knowledge.

---

[1]Code available at https://github.com/prakharguptaz/EDGE-exemplars

By using semantic frames from exemplars, EDGE outperforms a set of generative and retrieval-based baselines in a quantitative evaluation of response quality (coherence, consistency, fluency and diversity of responses), and outperforms token-based approaches in capturing the semantic structure of exemplar responses. Experiments demonstrate that semantic frames capture the meaning of the exemplars rather than their surface forms, such that EDGE does not copy inappropriate tokens from the exemplars. In a zero-shot anti-scam application, we show that EDGE generates exemplar-conditioned responses that are coherent, context-specific, and adherent to underlying exemplar intents and their high-level goals. To our knowledge, this work is the first to use frame semantics as a means of control in exemplar-based dialogue generation.

## 2 Frame Semantics

To achieve fluent and contextually-appropriate generated responses that adhere to the semantic structure of exemplars and capture their high-level goals, we use the *frame semantics* of the exemplars to guide the generation of responses. The central idea of frame semantics is *frames*, which are semantic abstractions describing universal categories of events, concepts, and relationships, based on the linguistic resource FrameNet (Baker et al., 1998). Frame semantics provide a higher-level representation of individual tokens in the response exemplars based on the purpose of those tokens in the response. For instance, the tokens 'hear', 'say', 'see', 'smell', 'feel', all share a similar purpose of their semantic frame label 'Perception', such that each frame can have many possible lexical surface forms. FrameNet defines more than 1200 frames such as 'Perception'.

Representing response exemplars in terms of their semantic frames allows our model to reuse their semantic structure to adapt the low-level response tokens to fit novel dialogue contexts, and produce diverse response variations that fit within the semantic constraints. For example, in Table 1, EDGE generates multiple diverse and coherent variations for both exemplar responses by conditioning on their frame semantic structures.

The use of frame semantics to represent exemplars in terms of their semantic meaning rather than their surface forms provides two additional benefits: (1) preserving the semantic structure of exemplars helps to preserve implicit constraints of dialogue systems present in exemplar responses including desired strategies, intents, and emotional tones, and (2) using frames rather than tokens helps the model to avoid overfitting. A model that uses exemplar tokens rather than frames during training can become over-reliant on copying tokens, such that during generation the model copies inappropriate tokens from the exemplar response. For example, given the exemplar response "Eggs are very beneficial for your body" (Table 1), a token-based model can access the token "Eggs" and incorrectly use "Eggs" in its response about vegan food. EDGE reduces such overfitting by conditioning on the semantic frames of the exemplars during training and generation. For example, EDGE uses the frame "FOOD" as input instead of "Eggs" (Table 1), and substitutes an appropriate token ("Vegan food") in its generated response.

In our experiments, we find that using frame semantics in exemplar-conditioned dialogue generation improves the coherency of responses, while preserving the semantic structure and underlying intents of the exemplar responses.

## 3 Model

Our model EDGE extends a dialogue generation model TransferTransfo (Wolf et al., 2019) to control generation by including semantic frames from an exemplar response in addition to the dialogue history. TransferTransfo is based on the transformer architecture and fine-tunes a generative pre-trained model (GPT) (Radford, 2018) with two objective functions: (1) a language modelling objective, and (2) a next-utterance classification objective. The language modelling objective function maximizes the likelihood for a given sequence of tokens, and the next-utterance classification objective distinguishes a correct response for an input dialogue context from a set of randomly selected distractor responses. We adapt the TransferTransfo model to our setting by first replacing GPT with GPT-2 (Radford et al., 2019) as our base architecture. GPT-2 can be substituted with other language models such as Transformer-XL (Dai et al., 2019) or dialogue specific models such as DialoGPT (Zhang et al., 2020). To incorporate semantic frames from exemplar responses in the TransferTransfo architecture, we uniquely add tokens representing the semantic frames to the input context. Specifically, we concatenate the input context, a <bof> token, semantic frame tokens, a <bor> token, and the response (Figure 1). Prior work also uses concatenation to add different signals to the input for training dialog systems (Budzianowski and Vulić, 2019). Following TransferTransfo model, we also add token, position, and speaker role embeddings. For frame extraction from exemplars, we use the open-sesame model Swayamdipta et al. (2017) and their open-sourced implementation[2]. We use the frame predicates and ignore the arguments. Because there are no frames corresponding to wh-question words such as 'why' and 'how', 'yes' and 'no', question mark or pronouns, we add each of these tokens in the frame vocabulary.

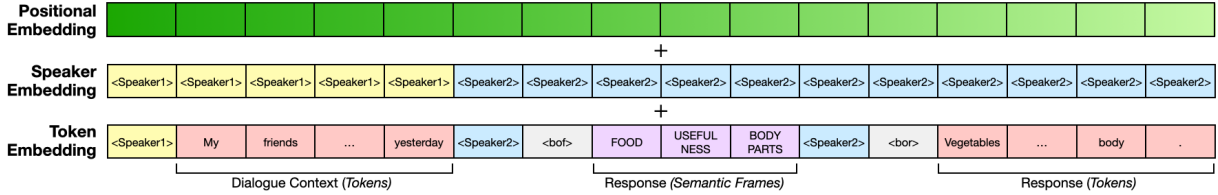

Figure 1: The input representation of our proposed approach. During training, EDGE conditions on the dialogue context and a noisy version of the ground truth response semantic frames to generate the ground truth response. During inference, we feed the context and the semantic frames from the response exemplars to generate a response.

**Training** During training, the model learns to generate the ground truth responses conditioned on the dialogue context tokens followed by the in-order predicted semantic frames for the ground truth response (Figure 1). Following TransferTransfo, we mask the tokens of the context for the language modelling objective. To ensure that the model does not ignore the exemplar response, we use the frames of the ground truth response in input during training, instead of frames from a retrieved response. In pilot experiments, our model generated incoherent replies to the dialogue context when the semantic frames were incorrectly detected or irrelevant to the dialogue context. To make the model more robust to missing frames, frames changing order between the exemplar and the response, and irrelevant or inaccurate frames, we: (1) randomly drop 15% of semantic frames from the sequence, (2) randomly shuffle semantic frames sequences (over a length of 2 tokens) with a probability of 0.1, and (3) add random semantic frames in random positions to increase the sequence length by 30%.

EDGE's ability to generate coherent responses despite inaccurate frame detection is important as the semantic frame prediction model that EDGE uses reports F1 scores of 73.25% for frame target detection and 86.55% for frame identification. However, informal dialogue text can lead to lower performance. Evaluating on 110 conversational sentences in the FrameNet 1.7 test set, the semantic frame prediction model achieves F1 scores of 71.78% for frame target detection and 74.58% for frame identification. We train EDGE by dropping, reordering and adding random frames so that EDGE learns to generate coherent responses in the presence of noisy frames from the exemplars.

**Inference** During inference, we either rely on predefined response exemplars, or perform retrieval by first using the state-of-the-art Poly-encoder model (Humeau et al., 2020) to retrieve response candidates and then select the highest ranked response as the exemplar response. We add the semantic frame sequence from the exemplar response as the input along with the context of the conversation. The model then creates a response which is controlled by the semantic frames from the exemplar, and coherent with the context of the conversation.

## 4 Experimental Setup

We compared our model to existing generative and retrieval-based approaches in two settings: (1) open-domain dialogue generation using the Dailydialog dataset (Li et al., 2017), and (2) goal-oriented anti-scam dialogue generation using a set of fraudulent emails (Radev, 2008) as prompts and a small set of intent-specific anti-scam response exemplars to inform responses. For the anti-scam domain, we investigated exemplar conditioned responses in a case without domain-specific training (i.e. zero-shot generation).

### 4.1 Datasets

**Open-Domain** We use the Dailydialog dataset (Li et al., 2017), which consists of 13,118 daily conversations covering topics such as culture, education, tourism and health. The validation and test sets have 1000 conversations each. We consider maximum of up to previous 5 utterances from the con-versation history as the context for both retrieval and generation. The 1000 conversations in the test set consists of 6740 such context-response pairs.

---

[2]https://github.com/swabhs/open-sesame

**Anti-Scam** We use fraudulent e-mails[3] as test data (Radev, 2008) consisting of 2500 emails. The intent of the fraudulent email sender (a scammer) is to convince the recipient to give the sender a large amount of money or some other information. We remove all links and email addresses from the email text and limit the text content to the first and last 3 sentences of the email, as these sentences typically reflect the setup and intent of the email, and the shorter email length reduces inference time.

### 4.2 Baselines

We compared EDGE with a set of baseline models:
- **Retrieval** (Humeau et al., 2020) The Polyencoder retrieval model allows for fast real-time inference by precomputing each candidate response representation once, and then ranking candidate responses for retrieval by attending to the context. Specifically, the model encodes two separate transformers, one for the context and one for the response, and creates multiple vector representations from the context. We use ParlAI's implementation[4] of this pre-trained transformer-based model.
- **GPT2-Gen** (Wolf et al., 2019) The dialogue generation model TransferTransfo (except that we replaced GPT with GPT-2). This model is the base architecture in our model. It uses the dialogue context to inform response generation, and does not condition on exemplar responses.
- **LSTM-Tokens** (Cai et al., 2019b) The state-of-the-art exemplar-conditioned open-domain response generation model. It uses the dialogue context along with tokens extracted from an exemplar response (using a transformer-based matching framework) to inform generation. LSTM with attention is used as the decoder.
- **LSTM-Frames** An ablation model that varies LSTM-Tokens to use the semantic frames from exemplar responses instead of extracted tokens. LSTM with attention is used as the decoder.
- **GPT2-Tokens** An ablation model that modifies EDGE to use tokens extracted from the exemplar response, as in (Cai et al., 2019b), instead of semantic frames. GPT-2 is used as the decoder.
- **GPT2-Frames (EDGE)** Our model that uses the dialogue context along with the semantic frames of the exemplar response to inform response generation. GPT-2 is used as the decoder.
- **Human** We collected human written responses for the test contexts.

We fine-tuned or trained each model on the Dailydialog dataset (Li et al., 2017).

### 4.3 Implementation Details

We use the architecture described in (Wolf et al., 2019) and use their open-source implementation with fine-tunable GPT-2 architecture[5]. We chose the 124M version of GPT-2 due to its performance and smaller size which accomodates resource constraints. We used the Adam optimizer with learning rate of 6.25e-5, L2 weight decay of 0.01, and batch size of 2. We set the number of candidates to 2 for the next-utterance classification objective. Each model was trained until maximum of 10 epochs with early stopping criteria. We set the maximum decoding length to 50 tokens and minimum to 4 for all models and use nucleus sampling (Holtzman et al., 2020) with threshold of 0.9. For LSTM-Tokens model, we used the open-sourced implementation released by the authors[6].

## 5 Results and Discussion

In this section we report results for both open-domain and goal-oriented anti-scam domains.

### 5.1 Open-Domain Setting

We compared EDGE with the baseline models on open-domain conversations in Dailydialog dataset, and report results in terms of human-rated and automatic metrics that capture aspects of response quality individually (*e.g.*, is the response grammatically correct?) and with respect to the context (*e.g.*, is the response a valid continuation of the preceding conversation?). We additionally consider how well the responses adhere to the semantic structure of the retrieved response exemplars.

### 5.1.1 Evaluation Metrics

Word overlap metrics have been shown to correlate poorly with human judgements of quality of responses (Liu et al., 2016) as they don't account for all the plausible responses for any given conversational context (Gupta et al., 2019). We therefore conducted human evaluations to capture aspects of the model quality such as coherence and fluency. Annotators on Amazon Mechanical Turk platform rated the responses of the models for 100 randomly selected test contexts on a scale of 1 to 3 (with 1 as the lowest and 3 the highest) on the following criteria:

- **Coherent** Does the response serve as a valid continuation of the preceding conversation?
- **Interesting** Is the response dull or interesting?
- **Fluent** Is the response naturally written, grammatical correct and non-repetitive?
- **Consistent** Does the response make logical sense given the context and by itself?
- **Uses semantics** Does the response share similar concepts with the retrieved response?

---
[3] https://kaggle.com/rtatman/fraudulent-email-corpus
[4] https://parl.ai/projects/polyencoder
[5] http://github.com/huggingface/transfer-learning-conv-ai
[6] https://github.com/jcyk/seqgen/tree/master/ranker

| Model | Dist-2 | Dist-3 | MaUdE | Coherent | Fluent | Consistent | Interesting | Semantics |
|---|---|---|---|---|---|---|---|---|
| Retrieval | 0.294 | 0.526 | 0.921 | 2.41 | 2.61 | 2.48 | 2.32 | - |
| GPT2-Gen | 0.249 | 0.494 | 0.905 | 2.42 | 2.55 | 2.41* | 2.18* | - |
| LSTM-Tokens | 0.182 | 0.380 | 0.890 | 2.04* | 2.10* | 2.11* | 1.89* | 2.17 |
| LSTM-Frames | 0.185 | 0.392 | 0.901 | 2.36* | 2.30* | 2.33* | 1.97* | **2.29** |
| GPT2-Tokens | 0.254 | 0.513 | **0.927** | 2.19* | 2.47* | 2.29* | 2.11* | 2.04* |
| EDGE (Ours) | **0.278** | **0.571** | 0.922 | **2.52** | **2.63** | **2.56** | **2.39** | 2.24 |
| Human | 0.385 | 0.720 | 0.911 | 2.76 | 2.69 | 2.78 | 2.44 | - |

Table 2: Results for automatic (Dist-2, Dist-3, and MaUdE) and human (Coherent, Fluent, Consistent, Interesting, and Uses Semantics) evaluation on the Dailydialog corpus. Our model significantly outperforms other models (t-test comparison with EDGE, $p < 0.05$ indicated with *) on human-rated metrics and performs similarly to the Retrieval baseline and Ablation models in automatic metrics. We did not collect Uses Semantics for the Human, Retrieval and GPT2-Gen cases which do not condition on exemplars.

The annotators were shown a conversational context and responses to rate, and were provided more detailed instructions and examples for each criteria, following Mehri and Eskenazi (2020). We collected ratings from 3 workers per context for all 7 models, with a total of 2100 ratings. The Cohen's Kappa (Cohen, 1968) value for inter-annotator agreement is 0.45 for the annotations, indicating moderate agreement. We also evaluate the models using an unreferenced automated evaluation metric **MaUdE** (Sinha et al., 2020) which uses large pre-trained language models to extract latent representations of utterances and is trained using Noise Contrastive Estimation. It has shown high correlation with human judgements on criteria such as interestingness and fluency. For measuring diversity of responses we calculate **Dist-n** (Li et al., 2016). It is the ratio of distinct n-grams to total number n-grams for all the responses from a model.

### 5.1.2 Results

The human evaluations in Table 2 demonstrate that (1) Unsurprisingly, the GPT-2 based models

| Metric | 1 Exemplar | 5 Exemplars | 10 Exemplars |
|---|---|---|---|
| GPT2-Gen | | | |
| Dist-2 | 0.240 | 0.129 | 0.096 |
| Dist-3 | 0.481 | 0.327 | 0.270 |
| LSTM-Tokens | | | |
| SemCov | 0.347 | 0.354 | 0.360 |
| Avg BLEU-2 | 0.216 | 0.214 | 0.214 |
| Dist-2 | 0.184 | 0.104 | 0.080 |
| Dist-3 | 0.387 | 0.267 | 0.223 |
| EDGE | | | |
| SemCov | **0.650** | **0.620** | **0.625** |
| Avg BLEU-2 | 0.192 | 0.170 | 0.161 |
| Dist-2 | **0.274** | **0.155** | **0.118** |
| Dist-3 | **0.569** | **0.409** | **0.344** |

Table 3: EDGE shows higher semantic coverage (SemCov) with the exemplar responses while showing lower lexical overlap (lower Avg BLEU-2). EDGE also achieves higher diversity (Dist-2,3).

(EDGE, GPT2-Tokens, and GPT2-Gen) achieve higher ratings for quality metrics of coherence, fluency, consistency, and interestingness compared to the LSTM based models (LSTM-Tokens and LSTM-Frames), and (2) The models that use semantic frames from retrieved responses (EDGE and LSTM-Frames) achieve higher ratings than the models that directly used tokens from the retrieved response (GPT2-Tokens and LSTM-Tokens). EDGE, our GPT-2 based approach that uses semantic frames from response exemplars, outperforms all other models on overall quality metrics, and outperforms token-based approaches in preserving semantics from reference responses. Both LSTM-Frames and EDGE achieve high Uses Semantics rating, indicating that the models which condition on frames preserve exemplar semantics better. EDGE and GPT2-Tokens also achieve the highest MaUdE scores as well as the highest Dist-n scores, indicating high quality and diversity of the responses.

| | | |
|---|---|---|
| **Context** | *Human1*: they sell everything. <br> *Human2*: well, i want chinese food. | *Human1*: actually i have a passion for chinese literature. <br> *Human2*: you do? |
| **Retrieved** | well, what do you want to eat? | yes, reading is my hobby. |
| **Frames** | WHAT DESIRING INGESTION ? | YES LINGUISTIC-MEANING |
| **GPT2-Gen** | it's a good idea. | yes. i'm passionate. |
| **LSTM-Tokens** | well, what's the you do? | yes, i do. |
| **LSTM-Frames** | i hope so. | yes, i did. |
| **GPT2-Tokens** | i'm not sure what to get. | what are you interested in? |
| **EDGE (Ours)** | you want to eat something chinese? | yes. i studied chinese literature at university. |
| **Context** | *Human1*: jeff, i'm going to the supermarket. do you want to come with me? <br> *Human2*: i think the supermarket is closed now. | *Human1*: did you go to the concert last weekend? <br> *Human2*: no, i didn't. and you? was it good? |
| **Retrieved** | i know. i intent to go to the store today. | yes, i did. i enjoyed it a lot. there was a folk singer, a violinist and a pianist. |
| **Frames** | AWARENESS PURPOSE MOTION BUSINESSES TEMPORAL-COLLOCATION | YES EXPERIENCER-FOCUS DESTINY LOCATIVE -RELATION PEOPLE |
| **GPT2-Gen** | what a pity! | yes. i enjoyed it very much. |
| **LSTM-Tokens** | yes, i'm sorry to go with you. | yes, i did. i've got a singer, but i was the violinist. |
| **LSTM-Frames** | where is the market? | yes, i've been interested in a lot of people. |
| **GPT2-Tokens** | where is the supermarket? | i think you're right. the performance was very beautiful. |
| **EDGE (Ours)** | i know, but i'm planning to go to the bank today. | yes. i was very interested in the performance. i was in the audience and it was really packed. |

Table 4: Sample model responses to dialogue contexts in the open-domain setting of Dailydialog conversations. The responses of all models except GPT2-Gen are conditioned on the Retrieved responses using either the retrieved response tokens or the extracted semantic frames (Frames). EDGE generates more coherent and interesting responses compared to the baselines, without directly copying tokens from the retrieved responses.

### 5.1.3 Semantic Coverage and Diversity

Our results demonstrate that EDGE generates higher-quality responses while preserving retrieved response semantics as rated by humans (Table 2). We further evaluate EDGE and baseline models (LSTM-Tokens, GPT2-Gen) to assess generated responses' consistency with retrieved responses, and the diversity of the generated responses (Table 3). We do not limit this experiment to the top retrieved response and instead select subsets of retrieved responses (of sizes 1, 5 and 10) for each test dialogue context by consecutively selecting each next highest ranked response if the maximum Jaccard similarity of its semantic frames with the semantic frames of any response in the subset is less than 0.5, and generate responses based on each response in the subset.

We calculate Dist-n to measure diversity, or the ratio of distinct to total n-grams for all the responses. EDGE achieves higher diversity than LSTM-Tokens and GPT2-Gen for all response set sizes. Compared to LSTM-Tokens, EDGE generated responses with semantic frames that covered a higher percentage of the semantic frames present in the retrieved responses (SemCov is 36% for LSTM-Tokens, and 63% for EDGE). This shows that compared to baselines, our model does not ignore the exemplar responses. It also copied exact tokens less often as EDGE generated responses contained a lower level of token similarity to retrieved responses (BLEU-2 of 0.21 for LSTM-Tokens and BLEU-2 of 0.16 for EDGE). This shows that while EDGE better controls the semantic content of the generated responses, it still produces more token-level diversity than other models (Dist-2, Dist-3).

### 5.1.4 Qualitative Analysis

We present sample dialogue contexts and model responses to demonstrate how EDGE performs on a range of retrieved response scenarios (Table 4). Overall, EDGE controls the length and semantic structure of its responses based on retrieved human-written exemplars, and thus produces longer and more specific responses compared to the purely generative model, GPT2-Gen. EDGE benefits from this exemplar-based control, even when retrieval or frame extraction fails. When the retrieved responses are not appropriate for the dialogue context (left two examples), EDGE leverages the semantic frames in the retrieved response to generate a coherent and specific response (e.g., by adding details such as "eat something chinese?"), while other models generate short or incoherent responses (e.g., "what a pity?"). When some words in the retrieved response are missing semantic frames (top right example), EDGE leverages the frames that are still present and the context to generate a coherent response with contextually-appropriate details. On the other hand, when LSTM-Tokens inappropriately copies tokens (top left and bottom right examples), the responses often become incoherent (e.g., copying "singer" and "violinist" results in "i've got a singer, but i was the violinist."). Although EDGE generates context specific responses which generally adhere to the semantics of the exemplars, EDGE still occasionally diverges from the exemplar response. For instance, the model can hallucinate details irrelevant to the context (the word "bank" in the bottom left example), a problem common in neural generative models (Tian et al., 2019; Dušek et al., 2020; Li et al., 2020).

| | | |
|---|---|---|
| **Context** | | i want you to assist in investing money... want to acquire stock in multi national companies and to engage in safe investments.... emphasize high level of confidentiality, which this upcoming project demands .... my son shall tell you where the funds are being maintained and discuss remuneration for your services |
| **Intent** Ask for Contact or Location | **Exemplar** | Are you located near me? Could we meet up soon about this? Tell me where your place is. |
| | **Frames** | LOCATING LOCATIVE-RELATION ? POSSIBILITY MAKE-ACQUAINTANCE TOPIC AWARENESS LOCATIVE-RELATION |
| | **Response** | would you please be able to meet me about the location of the funds? |
| **Intent** Show Interest | **Exemplar** | Very excited about the 20 million dollars you have promised me. I can use that for my business. |
| | **Frames** | DEGREE EMOTION-DIRECTED PROPORTIONAL-QUANTITY CARDINAL-NUMBERS POSSESSION COMMITMENT CAPABILITY USING BUSINESSES |
| | **Response** | i will be very happy to have the opportunity to use your company, as you have promised, i can use my company as an intermediary to invest in this business. |
| **Context** | | there is an account opened in this bank ... no body has operated on this account ... if i do not remit this money out urgently it would be fortified for nothing..... we will achieve this money together ... i will use my position in the bank to influence legal approvals and transfer of this money into your account with clearance from the ministries and foreign exchange department ... you will get 25% of the total amount. |
| **Intent** Show Skepticism | **Exemplar** | Why do you think I will give you any donation? I do not even know you. |
| | **Frames** | WHY INTENTIONALLY-ACT AWARENESS GIVING QUANTIFIED-MASS ? GIVING AWARENESS |
| | **Response** | why do you think i will give you such a large amount? i don't know you. |
| **Intent** Show Disinterest | **Exemplar** | I don't wish to undertake such an important duty. No one has ever requested me about this. |
| | **Frames** | INTENTIONALLY-ACT DESIRING ATTEMPT IMPORTANCE BEING-OBLIGATED REQUEST |
| | **Response** | i don't want to try to influence the decision of the ministry, the bank is not an important one and i don't want to interfere in the process. |

Table 5: Controlled response generation from EDGE in Anti-Scam domain. For each context (an initial scam email), we show two different intents and human-written response exemplars that embody each intent. We show each exemplar's extracted semantic frames, and EDGE's generated response. EDGE generates response variations by conditioning on exemplars to capture the specific exemplar intents.

## 5.2 Anti-Scam Setting

Traditional dialogue systems use response exemplars to control system responses based on high-level goals and intents present in the exemplar responses. However, it can be infeasible to write an exhaustive set of exemplar responses. Further, when such systems directly apply a pre-written response to a novel dialogue context, the response can be incoherent. We demonstrate an application of EDGE in the anti-scam domain where we generate a variety of coherent responses to novel dialogue contexts that capture the high-level intents of exemplar responses without training the models on domain-specific data (a zero-shot test scenario). We crafted our anti-scam response exemplars to follow high-level objectives of the domain (Dalton et al., 2020) such that each of our 20 response exemplars demonstrates one of 5 specific anti-scam intents: *ask for details, ask for contact or location, show interest, show skepticism,* and *show disinterest*. Half of the response exemplars contain generic replies that may be appropriate for many scam emails, and half of response exemplar replies contain responses to specific emails. We include sample scam emails, strategic response exemplars, and generated responses in Table 5.

| Model | Coherence | Intent | Engagement |
|---|---|---|---|
| GPT2-Gen | 2.10 | 33.0 | 70.1 |
| EDGE | **2.39** | **79.7** | **87.3** |

Table 6: Human evaluation of Coherence (reported from 1-low to 3-high), Intent (Follows Intent reported as a percentage), and Engagement (reported as a percentage) in the Anti-Scam setting.

**Human Evaluation** We performed human evaluation to test whether generated responses: (1) capture the high-level intents of the exemplar responses, and (2) generate coherent and engaging responses to the scam emails. We compared our system with the GPT2-Gen model, a GPT-2 based baseline that generates responses without conditioning on response exemplars. For each of the 20 response exemplars, we selected 5 scam emails as test dialogue contexts (100 emails total). We asked annotators to rate the responses of both models on the following criteria: (1) *Coherence*, or is the response on topic and strongly acknowledges the conversation history, (2) *Follows intent*, or does the response capture the intent of the exemplar, and (3) *Engagement*, or will the response engage the scammer in a conversation. We collected 3 ratings per email and averaged the ratings (Table 6) and the inter-annotator agreement (Cohen's Kappa) is 0.67 indicating high agreement. EDGE outperforms GPT2-Gen across all metrics, generating coherent replies that capture intents of the exemplars, and engage the scammer (high-level goals).

**Qualitative Analysis** GPT2-Gen responses often simply acknowledge the scammer's email (*e.g.*, "i am glad to tell you that i am in charge of your company." and "thank you, i'm sure you've got it" for the contexts in Table 5), while EDGE leverages the exemplars to generate longer replies that preserve the engagement aim and specific intent aims (*e.g.*, "i can use my company as an intermediary to invest in this business." to show interest). GPT2-Gen achieves 33% intent accuracy, even without conditioning on response exemplars, because its responses often showed interest or asked for details (two of the possible intents). While EDGE responses were more coherent, incoherent responses were typically due to long response exemplars, such that the resulting responses displayed faulty logic, a common problem across generative models generating long text (Holtzman et al., 2020). Overall, EDGE can leverage the semantic frames of response exemplars to preserve their underlying intent and add context specific details where appropriate (*e.g.*, "influence the decision of the ministry" in the last example). Thus, EDGE's key advantages over prior approaches are its controllability and zero-shot performance.

## 6 Related Work

EDGE controls dialogue generation based on semantic frames of exemplars, building on prior retrieval-based, controllable and semantics-based language generation methods.

**Retrieval-Based Generation** has been applied in summarization, machine translation, and paraphrasing (Peng et al., 2019; Gu et al., 2018; Grangier and Auli, 2018) tasks to improve the quality of text generation or to incorporate knowledge from retrieved text (Hua et al., 2019; Prabhumoye et al., 2019). In dialogue generation, retrieval conditioned approaches have been proposed to address the lack of diversity in generated responses and the generation of short and dull responses, common in generative approaches. Early approaches used LSTM-based models (Weston et al., 2018; Pandey et al., 2018; Wu et al., 2019) and their ensembles (Song et al., 2018; Zhang et al., 2019) to encode tokens of the retrieved responses to condition response generation. Conditioning response generation directly on tokens of retrieved responses results in: (1) generating incoherent responses due to copying contextually irrelevant tokens, and (2) models learning to ignore retrieved responses due to a mismatch between retrieved responses and ground truth responses. Prior work aimed to solve these problems by extracting only contextually relevant tokens from the retrieved response Cai et al. (2019a), and by replacing the retrieved response with a noisy version during training Cai et al. (2019b). By using semantic frames that represent an exemplar token's meaning rather than the low-level tokens themselves to guide generation, EDGE exerts better semantic control over the generated response. We additionally achieve higher coherence, fluency, and token-level diversity by reusing semantic frames rather than specific tokens.

**Controllable Text Generation** has been studied in tasks such as dialogue generation (Gao et al., 2019), summarization (Fan et al., 2018), paraphrasing (Goyal and Durrett, 2020), and other tasks (Dong et al., 2017; Peng et al., 2019), with the aim of controlling fixed attributes such as topic (Wang et al., 2017; Tang et al., 2019), emotion (Zhou et al., 2018), politeness (Niu and Bansal, 2018) and style (Keskar et al., 2019) through coarse-level labels or control phrases (Wu et al., 2020). Some traditional approaches used templates to control the generation of text (Reiter et al., 2005; McRoy et al., 2003). Some recent approaches learn templates from the data and exemplars (Wiseman et al., 2018; Ye et al., 2020; Yang et al., 2020). We explore the common case of response exemplars instead of inflexible templates or coarse labels to guide the dialogue response generation. Although state-of-the-art models pretrained on large dialogue corpus such as DialoGPT (Zhang et al., 2020), Meena (Adiwardana et al., 2020) and Blenderbot (Roller et al., 2020) are capable of generating interesting and human-like responses, our focus is on controlling the response generation process by conditioning on exemplars. By using semantic frames from exemplar responses, our method flexibly captures intents implicitly present in the exemplar frames, and exercises fine-grained semantic control over generation of new responses based on these exemplars.

**Semantics-Based Generation** has reemerged for use in various tasks such as paraphrasing (Wang et al., 2019), machine translation (Marcheggiani et al., 2018) and story generation (Tu et al., 2019; Fan et al., 2019). Semantic representations such as semantic frames and semantic role labels provide abstractions that capture the underlying meanings of different surface realizations (*e.g.*, paraphrases, other languages). We are the first to explicitly model frame semantic representations (Fillmore, 1982) in dialogue generation.

## 7 Conclusion

We present EDGE, an exemplar-based generative dialogue model. By generating responses that preserve semantic structures from exemplars, EDGE maintains desired qualities of dialogue systems including intents and strategies implicitly present in the curated exemplar sets, while achieving fluent and coherent responses. In future work, we plan to explore new mechanisms for incorporating semantic frames, experiment with other abstract representations of response exemplars, and apply our approach to other language generation tasks.

## 8 Acknowledgment

This work was funded by the Defense Advanced Research Planning Agency (DARPA) under DARPA Grant N6600198-18908, and the National Science Foundation under Awards No. IIS1816012 and IIS2007960.